\documentclass[10pt,twocolumn,letterpaper]{article}

\usepackage{cvpr}              

\usepackage{graphicx}
\usepackage{amsmath}
\usepackage{amssymb}
\usepackage{booktabs}
\usepackage[super]{nth}
\usepackage{adjustbox}
\usepackage{float}

\usepackage[pagebackref,breaklinks,colorlinks]{hyperref}

\usepackage[capitalize]{cleveref}
\crefname{section}{Sec.}{Secs.}
\Crefname{section}{Section}{Sections}
\Crefname{table}{Table}{Tables}
\crefname{table}{Tab.}{Tabs.}

\begin{document}

\title{OccTransformer: Improving BEVFormer for 3D camera-only occupancy prediction}

\author{
Jian Liu$^{1}$  \quad Sipeng Zhang$^{2}$ \quad Chuixin Kong$^{3}$ \quad Wenyuan Zhang$^{4}$ \quad Yuhang Wu$^{5}$ \\
Yikang Ding$^{6}$ \quad Borun Xu$^{3}$ \quad Ruibo Ming$^{6}$ \quad Donglai Wei$^{7}$ \quad Xianming Liu$^{1*}$  \\
$^{1}$HIT \quad $^{2}$ZJU   \quad $^{3}$UESTC \quad $^{4}$XJTU \quad $^{5}$HUST  \quad$^{6}$THU \quad$^{7}$FDU  \\
{\tt\small } 
}

\twocolumn[{
\vspace{-5em}
\maketitle
\vspace{-5em}
\begin{figure}[H]
\hsize=\textwidth 
\centering
\includegraphics[width=15cm]{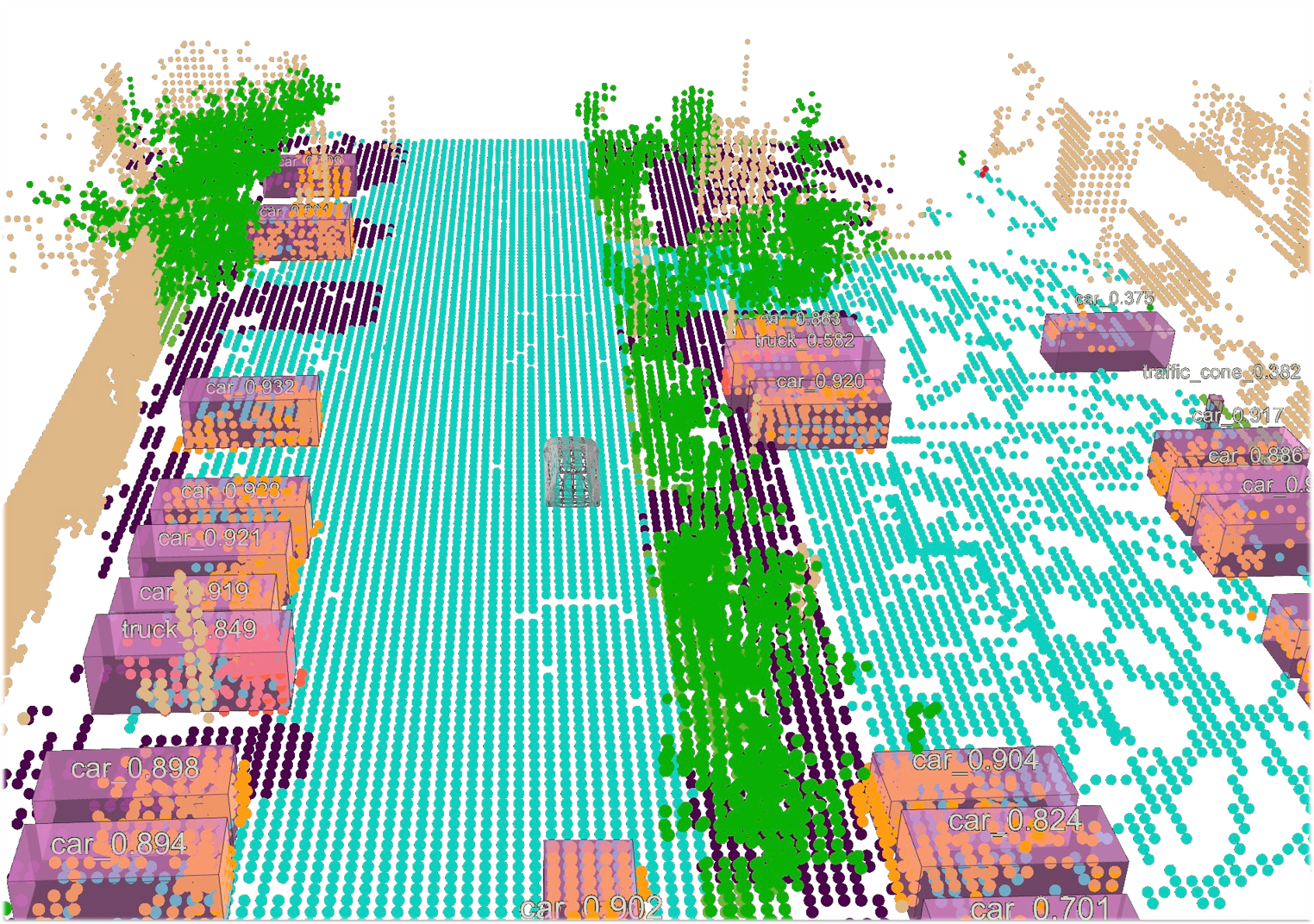}
\caption{Occ results on the testing set of nuScenes Dataset 3D Occupancy prediction track. BBox is 3D detection results}
\end{figure}
}]

\renewcommand{\thefootnote}{\fnsymbol{footnote}} 
\footnotetext[1]{~Corresponding author: Xianming Liu (csxm@hit.edu.cn)}

\begin{abstract}
This technical report presents our solution, "occTransformer," for the 3D occupancy prediction track in the autonomous driving challenge at CVPR 2023. Our method builds upon the strong baseline, BEVFormer ~\cite{li2022bevformer}, and improves its performance through several simple yet effective techniques. Firstly, we employed data augmentation to increase the diversity of the training data and improve the model's generalization ability. Secondly, we used a strong image backbone to extract more informative features from the input data. Thirdly, we incorporated a 3D Unet Head to better capture the spatial information of the scene. Fourthly, we added more loss functions to better optimize the model. Additionally, we used an ensemble approach with the occ model BevDet ~\cite{huang2022bevdet4d} and surroundOcc ~\cite{wei2023surroundocc} to further improve the performance. Most importantly, we integrated 3D detection model StreamPETR ~\cite{wang2023exploring} to enhance the model's ability to detect objects in the scene.
Using these methods, our solution achieved 49.23 miou on the 3D occupancy prediction track in the autonomous driving challenge. 
 
\end{abstract}

\section{Introduction}

The nuScenes image-based 3D occupancy prediction challenge at CVPR 2023, held from May \nth{12} to \nth{12}, 2023, is the largest and most exciting competition for challenging perception tasks in autonomous driving. This challenge was endorsed by Mobileye at CES 2023 and Tesla AI Day 2022, highlighting the importance of advancing perception technology for autonomous driving. The objective is to predict the current occupancy state and semantics of each voxel grid in the scene, given images from multiple cameras.

To tackle this challenge, our approach involves several steps to improve the accuracy and robustness of the model. 
Firstly, we enhance the baseline model by proposing a new model called occTransformer. Secondly, we use ensemble methods with multiple occupancy models to further improve the model's performance. Finally, we experiment with using a detection model to convert occupancy predictions and improve the model's mIoU for dynamic objects. By taking these steps, we aim to create a more effective and reliable model for 3D occupancy prediction.


\begin{figure*}
\begin{center}
\vspace{-5mm}
\includegraphics[width=0.85\textwidth]{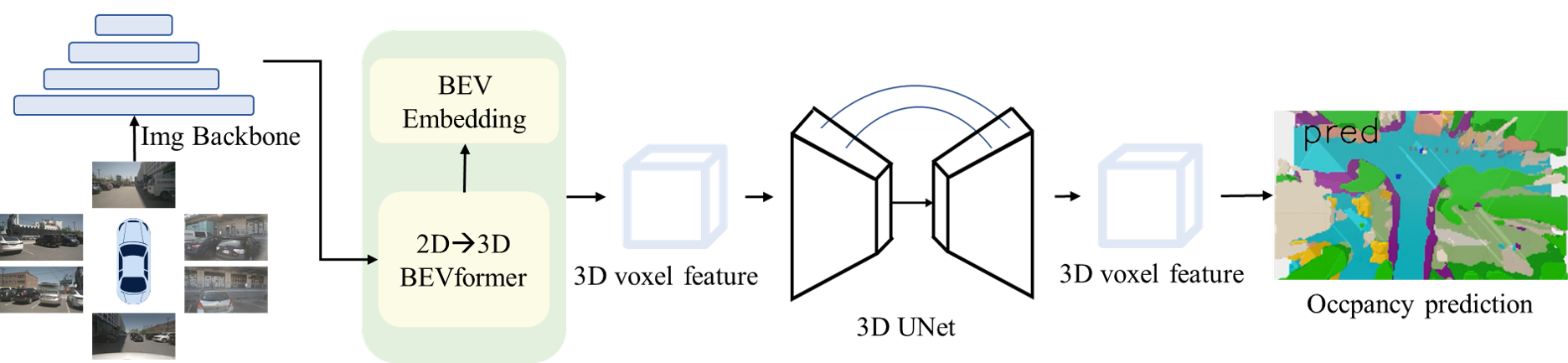}
\end{center}
\caption{The occTransformer framework involves the use of the bevformer method. These 2D features are first extracted and then aggregated into a bird's eye view (BEV) embedding. A simple decoder is then used to generate 3D voxel features, which are further enriched using a 3D U-Net head. The final output of the framework is a 3D occupancy prediction.}
\label{fig:model pipeline}
\end{figure*}

\section{Methods}

3D Occupancy Prediction aims to predict a 3D semantic scene with multi-camera images as input. The occupancy prediction is represented as dense cubic features $V \in \mathbb{R}^{H \times W \times Z \times C}$, where $ H, W, Z$ are the spatial resolution of the voxel space, $C$ denotes the semantic label of each voxel.

In this section, we present our solution to the 3D occupancy prediction challenge. We first introduce our data augmentation, followed by a detailed explanation of our models. Finally, we describe the ensemble strategy we used to achieve our results.

\subsection{data augmentation}
\textbf{Input and Data Augmentation.}To encourage the model to rely on more local features, which can help in training more robust and resilient models,  we apply the cutout augmentation ~\cite{devries2017improved} on the input multi-camera images $I=\{I^1,I^2,...I^N\}$. Cutout is a widely used image data augmentation that randomly selects certain regions within an input image and masks them by setting their pixel values to zero (or some fixed value), and we do not use any other augmentations in our method.

\subsection{Model}

We use BEVFormer~\cite{li2022bevformer} as  our baseline model, and several significant improvements were proposed during the challenge, which are introduced as follows.

\textbf{2D to 3D.} In this study, we aimed to improve the 3D perception tasks by replacing the baseline's bev generation module with different methods, including LSS~\cite{philion2020lift}, Flosp~\cite{cao2022monoscene}, and Flosp Depth~\cite{miao2023occdepth}. The lifting of 2D to 3D is a crucial aspect of these tasks. However, our experiments, as shown in Table ~\ref{tbl:2d3d}, revealed that the 2D-3D module proposed by BEVFormer~\cite{li2022bevformer} is the most effective method. It is important to note that all experiments were conducted without camera mask.

\textbf{Image Backbone.} Given a set of surrounding multi-camera images, the image backbone(e.g. ResNet-101~\cite{he2016deep}) extracts multi-scale image features. To obtain more detailed visual clues, we use Swin-L~\cite{liu2021Swin}, InternImage-XL~\cite{wang2023internimage}, convnextv2-L~\cite{woo2023convnext} as image backbone, respectively.

\textbf{3D Occupancy Head.} As shown in Figure~\ref{fig:model pipeline}, after lifting image feature from 2D to 3D, we can obtain the bev queries $F_{bev} \in \mathbb{R}^{H \times W \times C}$, where $C$ is embedding dims of each bev query. Then a two-layer MLP is utilized to decode 3D voxel feature $F_{3D} \in \mathbb{R}^{H\times W \times Z}$ from bev queries $F_{bev}$. Further more, following MLP, we use 3d UNet to get fine-grained 3D voxel feature. Specifically, 3d Unet module strengthens the spatial representation by fusing multi-scale voxel feature. In practice, we set the downscale ration = $\{2,4,8\}$. The output of 3d Unet $F_{unet}$ is used as occupancy predictor input.

\textbf{Loss}. The baseline model uses cross-entropy loss to supervise the occupancy prediction. We find that only using cross-entropy loss will result in ambiguous occupancy boundaries. To tackle this problem, we add dice loss~\cite{sudre2017generalised} when training. Thus, the overall loss is formulated as: 

$$L = \lambda_{ce} L_{ce}+ \lambda_{dice} L_{dice}$$




\subsection{Ensemble}
In this section, we describe our ensemble strategy. We found that weighting the probabilities from different models is more effective than taking the maximum probability or using a voting approach to combine the predictions from different models. Therefore, we used a weighted approach to combine the probabilities from different models.

\textbf{Pretrain Backbone Ensemble.} We first use different image backbone to train several improved version of BEVFormer models. 

\textbf{Occ Model Ensemble.} In addition to our own ensemble strategy, we also reproduced other methods for ensembling, including BEVDet ~\cite{huang2022bevdet4d}, SurroundOcc ~\cite{wei2023surroundocc}, and VoxFormer ~\cite{li2023voxformer}.

\textbf{DET Ensemble.} We found that 3D object detection models sometimes perform better than occ models on dynamic classes, so we decided to incorporate the detection model into our approach. To do this, we used StreamPETR ~\cite{wang2023exploring} to generate bbox frames and convert them into 3D bbox occ results. The process involved setting a threshold for each class based on the score and selecting high-confidence boxes. Then, we generated point clouds with a spacing of t within each box and checked whether each point was inside the box. After voxelizing the points, we assigned the corresponding semantic label to the voxels inside the box. When multiple semantic labels were predicted for a voxel, we selected the one with the highest score. Finally, we obtained the occ result and performed a prob average ensemble with the previous occ model.

\begin{table}[t]
\begin{center}
\vspace{4pt}
\resizebox{0.25\textwidth}{!}{%
\setlength\tabcolsep{6pt}
\begin{tabular}{l|c}
\hline
2d to 3d method   &miou   \\
\hline\hline
BevFormer ~\cite{li2022bevformer}  &23.464 \\
Flosp Depth ~\cite{miao2023occdepth}   &23.18\\
Flosp ~\cite{cao2022monoscene}       &22.038 \\
Lss  ~\cite{philion2020lift}    &22.773 \\
\hline
\end{tabular}
}
\end{center}
\caption{2d to 3d method w/o camera mask} 
\label{tbl:2d3d}
\end{table}

\begin{table}[!t]
\centering
\setlength\tabcolsep{2pt}
\begin{tabular}{c|c|c|c}
\hline
id & Method & image backbone & mIOU \\
\hline
\hline
a & Bevformer & res101~\cite{he2016deep} & 40.6 \\
b & Bevformer & swin-L~\cite{liu2021Swin} & 42.9 \\
c & Bevformer & convnextv2-L~\cite{woo2023convnext} & 44.0 \\
d & Bevformer & InternImage-XL~\cite{wang2023internimage} & 43.3 \\
\hline
e & Bevdet & swin-B~\cite{liu2021Swin} & 43.1 \\
f & SurroundOcc & InternImage-B~\cite{wang2023internimage} & 40.7 \\
g & VoxFormer & res101 ~\cite{he2016deep} & 40.7\\
\hline
h & ensemble b+c+d &- & 46.28 \\
i & ensemble h+e+f+g &- & 48.01 \\
j & ensemble i+det &- & 48.91 \\
\end{tabular}
\caption{Ensemble results.}
\label{tbl:ensemble_results}
\end{table}

\begin{table}[!ht]
    \centering
    \begin{tabular}{c c c c|c}
    \hline
        cutout & unet3d & dice loss & history frames & mIOU  \\ \hline
        -  &\checkmark  &\checkmark  &\checkmark  &37.7701  \\ 
        \checkmark  &-  &\checkmark  &\checkmark &37.5975  \\
        \checkmark  &\checkmark  &-  &\checkmark &34.5992  \\
        \checkmark  &\checkmark  &\checkmark &- &37.4897  \\ 
         \checkmark &\checkmark &\checkmark &\checkmark   &\textbf{38.1035}  \\
        
    \hline
    \end{tabular}
    \caption{Ablation study of our proposed improvements}
    \label{tbl:ablation study}
\end{table}


\begin{table*}
\vspace{-5mm}
\setlength{\tabcolsep}{0.005\linewidth}
\newcommand{\classfreq}[1]{{~\tiny(\semkitfreq{#1}\%)}}  

\def\mystrut{\rule{0pt}{1.5\normalbaselineskip}}
\centering
\begin{adjustbox}{width=1.99\columnwidth,center}
\begin{tabular}{l| c | c c c c c c c c c c c c c c c c c}
    \toprule
    Method 
    & \rotatebox{90}{mIou}
    & \rotatebox{90}{others} 
    & \rotatebox{90}{barrier}
    & \rotatebox{90}{bicycle} 
    & \rotatebox{90}{bus} 
    & \rotatebox{90}{car} 
    & \rotatebox{90}{construction vehicle} 
    & \rotatebox{90}{motorcycle} 
    & \rotatebox{90}{pedestrian} 
    & \rotatebox{90}{traffic cone} 
    & \rotatebox{90}{trailer} 
    & \rotatebox{90}{truck} 
    & \rotatebox{90}{driveable surface} 
    & \rotatebox{90}{other flat} 
    & \rotatebox{90}{sidewalk} 
    & \rotatebox{90}{terrain} 
    & \rotatebox{90}{manmade} 
    & \rotatebox{90}{vegetation}  \\
    \midrule
    NVOCC &54.19 &28.95	&57.98	&46.40	&52.36	&63.07	&35.68	&48.81	&42.98	&41.75	&60.82	&49.56	&87.29	&58.29	&65.93	&63.30	&64.28	&53.76\\
    42dot &52.45	&27.80	&56.28	&42.62	&50.27	&61.01	&35.41	&47.97	&38.90	&40.29	&56.66	&47.03	&86.96	&57.48	&63.64	&62.53	&63.00	&53.74\\
    UniOcc &51.27	&26.94	&56.17	&39.55	&49.40	&60.42	&35.51	&44.77	&42.96	&38.45	&59.33	&45.90	&83.90 &53.53	&59.45	&56.58	&63.82	&54.98\\
    occ-heiheihei  &49.36	&28.43	&54.49	&39.04	&45.45	&59.15	&32.05	&43.46	&36.33	&40.72	&51.67	&43.73	&84.97	&57.03	&61.38	&56.95	&57.95	&46.26\\
    \bf{occTransformer(Ours)} &49.23	&26.91	&53.57	&39.53	&47.56	&59.54	&32.59	&44.34	&37.36	&37.28	&54.81	&44.70	&84.61	&55.15	&60.34	&56.35	&57.14	&45.04\\
\bottomrule
\end{tabular}
\end{adjustbox}
\vspace{-4pt}
\caption{Top five submissions of 3d occupancy prediction track} 
\label{tbl:finalResults}
\end{table*}

\section{Experiments}

\subsection{Dataset and Evaluation}

The nuScenes dataset ~\cite{tong2023scene} \cite{tian2023occ3d} is a large-scale dataset specifically designed for autonomous driving research and has been fine-tuned for the 3D occupancy prediction competition. It consists of over 34,000 samples, including 28,130 samples for training, 6,019 samples for validation, and 6,008 samples for testing. The dataset includes data from six cameras and has a voxel size of 0.4m. The range of the dataset is from -40m to 40m in the x and y directions and from -1m to 5.4m in the z. The volume size is [200, 200, 16]. The dataset contains 18 classes, with classes 0 to 16 defined the same as in the nuScenes-lidarseg dataset. The label 17 category represents voxels that are not occupied by anything, which is named as \textit{free}. The ground truth labels of occupancy are derived from accumulative LiDAR scans with human annotations.

During training, both [mask lidar] and [mask camera] masks are optional, and participants are not required to predict the masks. However, during evaluation, only [mask camera] is used, and we use the provided camera mask as default.

\subsection{Implementation Details}

We use AdamW2~\cite{loshchilov2017decoupled} optimizer with cosine annealing policy. We set learning rate max to $2 \times 10^{-4}$, with 0.01 weight decay. The model is trained on 8 NVIDIA V100 GPUs with 24 epochs. We use R101-DCN as backbone during the exploration stage, and finally use Swin-L~\cite{liu2021Swin}, ConvNextv2-L~\cite{woo2023convnext}, InternImage-XL~\cite{wang2023internimage} to build stronger models.

Unless we explicitly indicate, all models used for ablation study, including Table~\ref{tbl:ensemble_results} and Table~\ref{tbl:ablation study}, are trained on training data, tested on validation data. In the final submission, we first train 24 epochs on training data and finetune 12 epochs on the whole trainval data. 

\subsection{Ablation Study}

Table \ref{tbl:ablation study} shows the ablation results of our improvements on baseline. We find that  the dice loss significantly boost the performance of occupancy. 

\section{Results}

In the 3D occupancy prediction track, we improved the BEVFormer model and effectively integrated the results of existing occupancy models. Finally, we utilized a detection model to improve the mIoU for dynamic objects. The top five final results are shown in Table ~\ref{tbl:finalResults}

\section{Acknowledgement}

We would like to express our sincere gratitude to Haotian Yao, Xiaoming Zhang, Jianming Hu, and Lihui Peng for their guidance and support throughout our research project.

{\small
\bibliographystyle{ieee_fullname}
\bibliography{egbib}
}

\end{document}